\documentclass[journal]{IEEEtran}

\usepackage{ifpdf}

\usepackage{cite}
\usepackage[makeroom]{cancel}
\usepackage{xcolor}
\usepackage{bm}
\usepackage{physics}

\ifCLASSINFOpdf
   \usepackage[pdftex]{graphicx}

\else

\fi

\usepackage{amsmath}
\usepackage{wasysym} 
\usepackage{amssymb}

\usepackage{algorithmic}
\usepackage{array}
\usepackage[caption=false,font=normalsize,labelfont=sf,textfont=sf]{subfig}

\usepackage{booktabs}
\usepackage{tabularx}
\newcolumntype{M}[1]{>{\centering\arraybackslash}m{#1}}
\newcolumntype{L}[1]{>{\raggedright\arraybackslash}m{#1}}
\usepackage{booktabs,caption} 
\usepackage[flushleft]{threeparttable}
\usepackage[makeroom]{cancel}
\usepackage{soul}
\usepackage{collect}
\usepackage{mathtools} 
\usepackage{empheq}
\usepackage{comment}
\begin{document}

\title{How to Model Brushless Electric Motors for the Design of Lightweight Robotic Systems}


\author{Ung Hee Lee$^{1,2,3}$,~\IEEEmembership{Student Member,~IEEE,}
Tor Shepherd$^{1,3}$,
Sangbae Kim$^{4}$,~\IEEEmembership{Member,~IEEE,}
Avik De$^{5}$,
\\Hao Su$^{6}$,~\IEEEmembership{Member,~IEEE,}
Robert Gregg$^{3,7}$,~\IEEEmembership{Senior Member,~IEEE,}
Luke Mooney$^{8}$,
Elliott Rouse$^{1,2,3}$,~\IEEEmembership{Senior Member,~IEEE}
\thanks{This work was graciously supported by the NSF National Robotics Initiative under award no. 1830338.}
\thanks{$^{1}$Neurobionics Lab, University of Michigan, Ann Arbor, MI 48109 USA}
\thanks{$^{2}$Department of Mechanical Engineering, University of Michigan}
\thanks{$^{3}$The Robotics Institute, University of Michigan}
\thanks{$^{4}$Department of Mechanical Engineering,
Massachusetts Institute of Technology, Cambridge, MA, 02139, USA}
\thanks{$^{5}$Department of Electrical and Systems Engineering, University of Pennsylvania, Philadelphia, PA 19104 USA}
\thanks{$^{6}$Lab of Biomechatronics and Intelligent Robotics, Department of Mechanical Engineering and Aerospace Engineering, North Carolina State University, Raleigh, NC 27695 USA}
\thanks{$^{7}$Department of Electrical Engineering and Computer Science, University of Michigan}
\thanks{$^{8}$Dephy Inc. Maynard, MA 01754 USA}
\thanks{Corresponding email: \tt\footnotesize ejrouse@umich.edu}
}

\markboth{Journal of \LaTeX\ Class Files,~Vol.~14, No.~8, August~2015}%
{Shell \MakeLowercase{\textit{et al.}}: Bare Demo of IEEEtran.cls for IEEE Journals}

\maketitle

\begin{abstract}

A key step in the development of lightweight, high performance robotic systems is the modeling and selection of permanent magnet brushless direct current (BLDC) electric motors. Typical modeling analyses are completed \textit{a priori}, and provide insight for properly sizing a motor for an application, specifying the required operating voltage and current, as well as assessing the thermal response and other design attributes \textit{e.g.} transmission ratio). However, to perform these modeling analyses, proper information about the motor's characteristics are needed, which are often obtained from manufacturer datasheets. Through our own experience and communications with manufacturers, we have noticed a lack of clarity and standardization in modeling BLDC motors, compounded by vague or inconsistent terminology used in motor datasheets. The purpose of this tutorial is to concisely describe the governing equations for BLDC motor analyses used in the design process, as well as highlight potential errors that can arise from incorrect usage. We present a power-invariant conversion from phase and line-to-line reference frames to a familiar q-axis DC motor representation, which provides a ``brushed'' analogue of a three phase BLDC motor that is convenient for analysis and design. We highlight potential errors including incorrect calculations of winding resistive heat loss, improper estimation of motor torque via the motor's torque constant, and incorrect estimation of the required bus voltage or resulting angular velocity limitations. A unified and condensed set of governing equations is available for designers in Appendix \ref{sec:summary}. The intent of this work is to provide a consolidated mathematical foundation for modeling BLDC motors that addresses existing confusion and fosters high performance designs of future robotic systems.

\end{abstract}

\begin{IEEEkeywords}
brushless electric motors, BLDC, design, modeling, field oriented control
\end{IEEEkeywords}

\IEEEpeerreviewmaketitle

\section{Introduction}

The success of numerous modern robotics and automation applications is predicated on the use of brushless electric motors. Electric motors convert electrical energy to and from mechanical energy, which requires coordinated interaction between electric and magnetic fields that produce torque via the Lorentz Law. Brushed electric motors use mechanical contacts---termed \textit{brushes}---to provide energy to the motor's winding that produces torque. More recently, brushless direct current (BLDC) motors have been developed to address challenges associated with mechanical brushes, where BLDC motors use multiple windings energized via electrical switching, a process known as \textit{electric commutation}. The absence of mechanical brushes enables BLDC motors to have improved efficiency, power density, longevity, and reduced audible noise \cite{Ozturk2010,neethu2012}. Consequently, BLDC motors are a compact, highly controllable, and relatively efficient actuation method popular across a wide range of robotics applications. \par

The electromechanical design process often requires careful assessment of motor technologies to ensure the motor is properly matched to the application. This is especially important for certain areas of robotics, such as legged robots, wearable robots, or autonomous vehicles, where actuator mass is a critical aspect of success. To ensure the motor is properly specified, the torque-velocity requirements of the application are translated to the required current, voltage, and thermal demands on the motor. These analyses are then used to make choices in the design process. For example, the required motor voltage and current govern the appropriate power supply, winding type, and transmission ratio; in addition, the expected temperature increase of the windings governs whether the motor is properly sized for the application. To complete these analyses during the design specification process, the parameters that govern BLDC motor operation (\textit{e.g.} torque constant, winding resistance, thermal resistance, \textit{etc.}) are required. \par

Motor datasheets provide critical information needed for analyses in the design specification process; however, different ways this information is provided can lead to common errors that ultimately lead to heavier designs. Roboticists usually simplify three phase BLDC motors to an analogous single phase ``brushed" motor, which is convenient and intuitive to assess, since it is described by a single current, voltage, resistance, inductance, torque constant, \textit{etc}. To this end, care must be used when reducing to the brushed electromechanical model. For example, the winding resistance provided in manufacturer datasheets is often the \textit{terminal resistance}, which is conveniently measured across two leads of the windings; however, terminal resistance should not be used with the q-axis current typically provided by commercial motor drives. When reducing properties of BLDC motors to a single-phase brushed analogue, the designed must ensure all parameters are considered in a consistent reference frame. For example, if improperly-matched resistance and current values are used in modeling or analysis,  estimates of heat produced (\textit{i.e.} Joule heating) can be over or under-estimated by as much as 100\%. Thus, clarity, standardization, and a strong mathematical foundation are needed to ensure BLDC motors are able to be conveniently and accurately modeled during the design process, especially when motor mass is a driving design factor.
\par
\begin{figure}[!t]
\centering 
 \includegraphics[width=0.8\columnwidth]{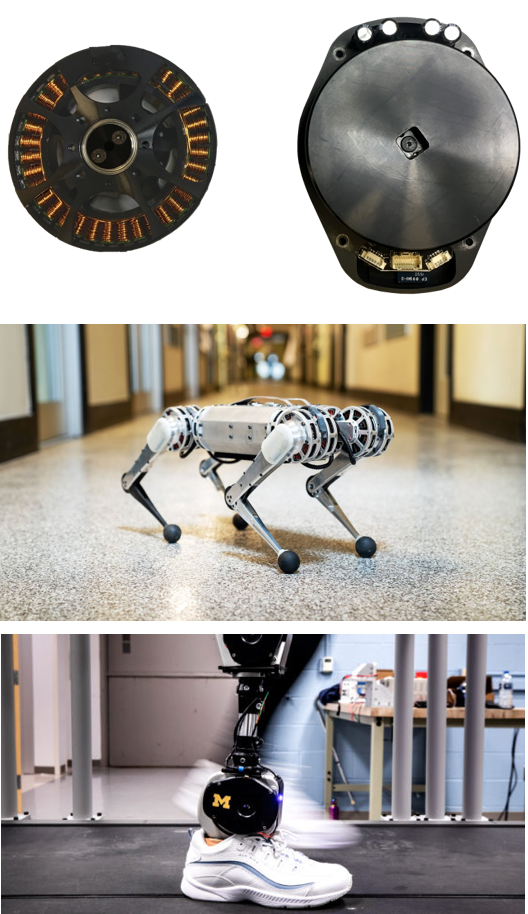}
\caption{The picture of the high-torque exterior-rotor type Brushless DC motor (T-Motor 100 KV, top left; Dephy Actuator Package, top right), and examples of robots using BLDC motors: MIT Cheetah Robot \cite{vickmark_2019} (middle), the University of Michigan's Open Source Bionic Leg \cite{jj_} (bottom). Figures used with permission.}
\label{figure:tmotor}
\end{figure}

We believe this tutorial will be a helpful resource to designers studying the use of high-performance BLDC motors, particularly in mobile robot applications (Fig. \ref{figure:tmotor}). In addition to the importance of motor modeling in the design process, other recent work has highlighted the need for greater standardization of analyses and terminology. This is evidenced by recent characterizations of new motors that do not have adequate manufacturer specifications \cite{lee2019empirical}. In addition, researchers have occasionally obtained parameters that differ from what is provided by motor manufacturers \cite{seok2013design}. Thus, an improved understanding of motor modeling and their underlying parameters will reduce the time and cost associated with actuator design, use, and characterization \cite{seok2013design,seok2014design,kenneally2016design, ding2017design,rezazadeh2018robot,katz2019mini,laschowski2019lower,torrealba2019toward, sarkisian2020design, elery2020design,yu2020quasi, azocar2020design}.
\par
There are many different types of brushless motor topologies, which can add to challenges modeling their performance. In the most general description of brushless motors, there are two types, namely BLDC motors and asynchronous electric motors (\textit{i.e.} AC induction motors). Within BLDC motors, further separation can be made based on the shape of their back electromotive force (back-EMF) profile as a function of angular displacement, being either sinusoidal or trapezoidally shaped. The waveform of the back-EMF profile stems from the distribution of each winding and the shape of stator slots and magnetic poles. The back-EMF shape affects the performance of motors and plays a role in the selection of an appropriate commutation paradigm to maximize efficient torque production \cite{Li2012}. Some previous works have denoted motors with trapezoidal back-EMF profiles as BLDC motors, with permanent magnet synchronous motors (PMSM) being used to denote motors with sinusoidal profiles \cite{neethu2012,lee2009comparison, pillay1991application}. This naming convention has become confounded by newly popularized ``drone" style exterior-rotor type motors being described as BLDC, despite having a sinusoidal back-EMF profiles. Thus, in this work, we do not distinguish between PMSM and BLDC motor types, and instead we limit our analysis to BLDC motors with a sinusoidal back-EMF profile for ease of analysis.
\par
In this tutorial, we describe the governing equations for brushless motor operation, and highlight inconsistencies that stem from misinterpretation of motor datasheets. We provide conversions from phase-based representations of motor current and voltage to a DC representation using the quadrature-direct transformation; by converting to a DC representation, the familiar and convenient single-phase (\textit{i.e.} brushed) DC motor analogue can be used for interpretation and design. We introduce three potential errors that arise when motor characteristics from datasheets are used without proper conversions, namely incorrect calculations of resistive losses, inaccurate estimation of motor torque via the torque constant, and improper determinations of maximum voltage needed for power supply specification. The intent of this work is to improve understanding of how brushless motors are described by manufacturer datasheets, and to enable accurate and high performance designs of future robotic systems.

\definecollection{mytable}
\newcounter{duptable}
\newcounter{tmp}

\makeatletter
\newcommand\resttable{%
\let\oldlabel\label
\let\label\@gobble
\setcounter{tmp}{\value{table}}
\setcounter{table}{\value{duptable}}
\includecollection{mytable}
\let\label\oldlabel
\setcounter{table}{\value{tmp}}
}
\makeatother

\begin{collect*}{mytable}{\setcounter{duptable}{\value{table}}}{}{}{}
\begin{table}[h]
    \renewcommand{\arraystretch}{1.3}
    \centering
    \caption{Nomenclature of electromechanical quantities}
    \begin{tabular}{L{1.25cm}L{1.0cm}L{5.25cm}}
    \hline
    Notation& Units &Description \\
    \hline 
    \vspace{0.05cm}
    $\bar{[\cdot ]}$&  & Amplitude of a sinusoidal quantity\\
    ${[\cdot ]}^\phi$&  & A quantity in the phase reference frame\\
    ${[\cdot ]}^{ll\text{ or } l}$&  & A quantity in the line-to-line (terminal) or line reference frame\\
    ${[\cdot ]}^q$&  & A quantity in the q-axis reference frame\\
    ${[\cdot ]}^{RMS}$&  & The root-mean-square of a phase quantity\\
     ${[\cdot ]}_{A,B,\text{ or } C}$&  & Specific phase of a BLDC winding\\
    $I$ & A & Current\\
    $I^a$ & A & Armature current\\
    $V$ & V & Voltage\\
    $V^{bus}$ & V & Bus voltage\\
    $V^{e}$ & V & Back-EMF voltage\\
    $R$ & $\Omega$ & Electrical resistance\\
    $L$ & H & Inductance\\
    $L^{e}$ & H & Effective inductance\\
    $K_t$& Nm/A & Torque constant \\
    $K_b$     & Vs/rad       & Back-EMF constant\\
    $K_v$     & rad/Vs       & Velocity constant\\
    $\tau $   & Nm         & Torque\\
    $P$         & W             & Resistive power loss \\
    $B$         &  T            & Magnetic flux density \\
    $F$         &  N            & Electromagnetic force \\
    $J $        & Kgm$^2$    & Rotor inertia \\
    $\theta_r$ & rad & Rotor angular displacement\\
    $\theta$ & rad & Magnetic angular displacement\\
    $b$         & Nm/rad/s              & Damping coefficient \\
    $\ell$         & m             & Length of a coil \\
    $r$         & m             & rotor radius \\
    $d$         & m             & Half width of a coil \\
    $N$         &              & Number of coils in the winding \\
    $p$         &               & Number of pole pairs \\
    $\bm j$         &               & Unit vector \\
    \hline 
    \addlinespace[0.05cm]
    \end{tabular}
    \begin{tablenotes}
    \item For example, $\bar{K_v^{ll}}$ is the line-to-line velocity constant, which represents the no load velocity divided by the amplitude of the sinusoidal line-to-line (terminal) voltage (sometimes known as a motor's $K_v$ number), is obtained by combining three different rows of the table.
    \end{tablenotes}
\end{table}
\end{collect*}
\section{Electromechanical Modeling of Motors} 
In this section, we describe the equations that govern brushed and BLDC motors, and introduce the concept of electrical commutation for BLDC motors. Most importantly, we present a brushed motor representation of the BLDC motor using direct-quadrature (d-q) transformation \cite{o2019geometric}. In this work, we make the following assumptions regarding BLDC motors:
\begin{itemize}
\item Non-salient\footnote{In this paper, we define saliency as magnetic saliency \cite{mevey2009sensorless,hendershot2010design}, which describes if a motor's winding inductances vary as a function of rotor angle. Note that this is different from salient poles, which stems from the physical characteristics of the magnetic poles. A motor could have salient poles but be a non-salient motor under this definition; typical hobbyist brushless motors fall under this category \cite{parsons_2018}.}
\item Three phase sinusodial back-EMF profiles
\item Identical mutual and self inductances
\item The d-q transformation is in phase with the rotor ($I^d = 0$)
\end{itemize}
These assumptions provide a more convenient analytical approach while also representing the real-world use case of BLDC motors. For example, an exterior rotor type BLDC motor (Fig. 1) satisfies these assumptions when used with an off-the-shelf field oriented control (FOC) drive and operating with a sufficient DC bus voltage / power supply  \cite{de2019task}.
\par
 In this paper, we define the \textit{winding reference frame} to collectively describe the different electrical references where superscript $\phi$, $l$, $ll$ and $q$ denote a phase, line, line-to-line, and q-axis quantities, respectively. \textbf{Bold font} denotes vectors or matrices, $\bar{bar}$ represents amplitude of a sinusoidal quantity, and we boxed the key equations of each section. All boxed equations are provided together in Appendix \ref{sec:summary}, meant to be an convenient reference guide for designers. Finally, some BLDC motors have multiple pole pairs, which causes a discrepancy between rotor angular displacement and magnetic angular displacement used in analyses and commutation. In this work, all angular displacements are represented in the magnetic domain, where one magnetic rotation is defined by a $1/p$ rotation of the rotor. To convert from magnetic displacement to rotor displacement, the following conversion can be used 
\begin{equation}
\theta = \theta_r \cdot p.
\end{equation}. 
\subsection{Brushed DC motors} \label{sec:brushed_modeling}
To accurately understand BLDC motors, we first describe the equations that govern ideal brushed DC motors in the electrical and mechanical domains. By applying Newton's Law, we can obtain
\begin{equation}  \label{eq:brushed_mech}
 \boxed{J\frac{d^2\theta_r}{d^2t} = K_t I^a -  b \frac{d\theta_r}{dt} - \tau_L}
\end{equation}
where $J$ is the inertia of the rotor, $K_t$ is the torque constant, $I^a$ is the armature current, $b$ is the viscous damping coefficient, $\theta_r$ and $\frac{d\theta_r}{dt}$ are the rotor angular displacement and the rotor's angular velocity, respectively, and $\tau_L$ is the load torque. Applying Kirchoff’s Voltage Law (KVL) across the winding yields:
\begin{equation} \label{eq:brushed_elec}
\boxed{V = RI^a + K_b \frac{d\theta_r}{dt} + L\frac{dI^a}{dt}}
\end{equation}
where $V$ is the voltage applied across the winding, $K_b$, is the back-EMF constant, $R$ is the resistance of the rotor winding, and $L$ is the inductance of the winding. The term $K_b \frac{d\theta_r}{dt}$ is known as the back electromotive force (back-EMF), which is the generated voltage that opposes the voltage across the winding ($V$). These two equations are the fundamental relationships that govern ideal brushed DC motor operation.

\subsection{BLDC motors} \label{sec:bldc_modeling}

In this section, we describe the governing equations that underlie the operation of ideal BLDC motors, beginning with a single phase, and expanding to three phase operation. Similar to the equations governing brushed DC motors, we summarize BLDC models for both the mechanical and electrical domains.

\subsubsection{Per-phase modeling}
\begin{figure}[t]
\centering 
 \includegraphics[width=\columnwidth]{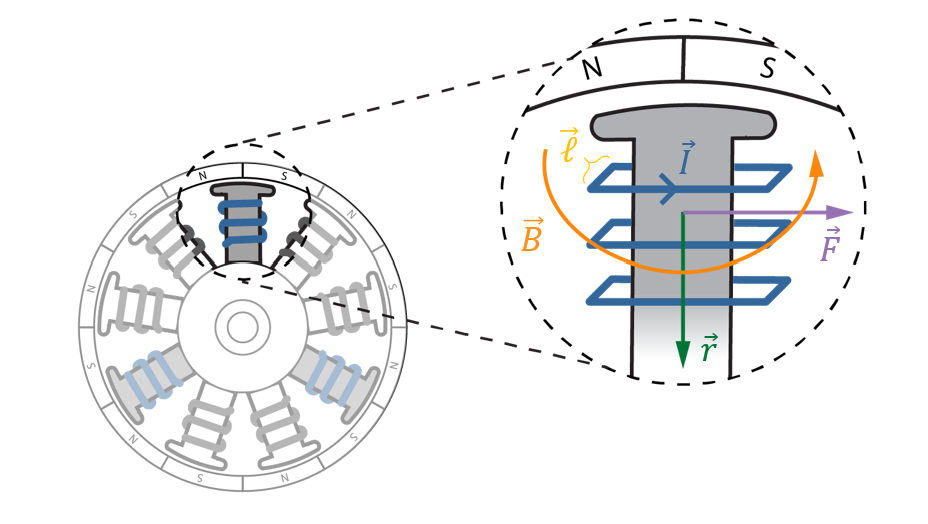}
\caption{A simplified depiction of an exterior-rotor type brushless motor shown with electromagentic force diagram. The gray colored poles indicates that they are part of the same phase. Note that for visualization purposes, electromechanical quantities are  depicted as a single pole value, although they are per-phase quantities (magnified view of the figure). }
\label{figure:motor_model}
\end{figure}

We first begin by understanding how a single phase produces torque. Per-phase torque production from a BLDC motor can be expressed as
\begin{equation} \label{eq:per-phase torque}
    \bm \tau^\phi = \bm F \cross \bm r = p( N  I^\phi \bm B^\phi (\theta) \cross \bm \ell ) \cross \bm r
\end{equation}
where $p$ is the number of magnetic pole pairs, $\theta$ is the angular displacement in the magnetic domain, $\phi$ denotes phase quantities, $\bm F$ is the electromagnetic force vector calculated from the Lorentz Law, $N$ is the number of individual coils in the winding, $\bm B$ is the per-phase magnetic flux density vector, and $\bm \ell$ is the vector length of a  coil side  along the axis  parallel to the rotation axis (perpendicular to the magnetic field), $\bm r$ is the rotor radius, and $I^\phi$ is the current flowing through one phase as a function of time (Fig \ref{figure:motor_model}).   \par

Since each winding is perpendicular to the magnetic flux density formed by the permanent magnets, the per-phase torque becomes: 
\begin{align}
\bm \tau^\phi &= pN I^\phi \norm{\bm { B^\phi (\theta)}} \norm{\bm \ell} \norm{\bm r} \bm j \;\;   ( \bm{ B^\phi}  (\theta)\perp \bm  \ell \perp \bm r) \\
\bm \tau^\phi& = pN A \norm{\bm{B^\phi (\theta)}}  \norm{\bm{I^\phi}} \bm j 
\end{align}
where $A = \ell r$ and $\bm j$ is a unit vector perpendicular to both the electromagnetic force and the rotor radius (vector into the page in Fig. \ref{figure:motor_model}), where $\norm{ \cdot }$ represents the Euclidean norm. 
As we assume the magnetic flux density of a BLDC motor is sinusoidal, the magnetic flux density can be expressed as a function of the magnetic angle of the rotor:

\begin{equation} \label{eq:magnetic_profile}
\norm{\bm{B^\phi (\theta)}} =- \bar{B^\phi}\sin(\theta)
\end{equation}
Therefore, the per-phase torque becomes:
\begin{equation}
  \bm \tau^\phi = -p N A  \bar{B^\phi}  I^\phi \sin(\theta) \bm j
\end{equation}
where the $\bar{B^\phi}$ is the amplitude of the per-phase magnetic field.

Eq. (\ref{eq:per-phase torque}) can be further simplified by defining the per-phase torque constant:
\begin{equation}
\boxed{\bm \tau^{\phi}= K^{\phi}_t  I^{\phi} \bm j}
\end{equation}
where the per-phase torque constant is defined as:
\begin{equation}
K^{\phi}_t(\theta) = - p N A \bar{B^{\phi}}  \sin({\theta}) 
\end{equation}
and the per-phase torque constant is a function of the rotor's position, where $\bar{K^\phi}(\theta)$ is the amplitude of the per-phase torque constant:
\begin{equation}
    K^\phi_t(\theta) = -\bar{K^\phi_t} \sin(\theta) 
\end{equation}
The sinusoidally varying per-phase torque constant highlights the difference from the brushed motor torque constant, which stems from the shape of the rotor magnetic flux density across rotor angle \cite{mevey2009sensorless}. \par
The per-phase back-EMF can be calculated using Faraday's Law:
\begin{equation}
\ V^{\phi}_e = N r (\bm B_{\phi} \cross \bm{ \frac{d\theta}{dt}}) \cdot \bm \ell
\end{equation}
where $V^{\phi}_e$ is the per-phase back-EMF. Since the magnetic flux density and radius of the coil are perpendicular, and the cross product of those are parallel to the vector length of the coil, the per-phase back-EMF equation reduces to:
\begin{equation}
 V^\phi_e = N A \norm{\bm {B^\phi (\theta)}} \norm{\bm{\frac{d\theta}{dt} }} \;\;   (\bm B^\phi(\theta) \cross \bm{ \frac{d\theta}{dt}}  \parallel \bm \ell)
\end{equation}
where the magnetic flux density has a sinusoidal profile as per (\ref{eq:magnetic_profile}): 
\begin{equation}
V^{\phi}_e = -N A  \bar{B_{\phi}}\sin({\theta} ) p \frac{d\theta_r}{dt}
\end{equation}
where $\frac{d\theta}{dt} = p \frac{d\theta_r}{dt}$, accounting for multiple pole pairs. By defining the back-EMF constant, the expression becomes: 
\begin{equation}
  V^\phi_e = - K^\phi_b (\theta) \frac{d\theta_r}{dt}
\end{equation}
where the per-phase back-EMF is the function of the rotor's position:
\begin{equation}
 K^\phi_b (\theta) =- p N A  \bar{B_{\phi}}\sin({\theta} )
\end{equation}
\begin{equation}
 K^\phi_b (\theta) = -\bar{K^\phi_b} \sin({\theta})
\end{equation}
The per-phase back-EMF profile is sinusoidal, similar to the per-phase torque constant, which also originates from the shape of the flux density. Note that ideally the per phase back-EMF is identical to the per-phase torque constant (\textit{i.e.} $K_b^\phi=K_t^\phi$).

\subsubsection{Three-phase modeling}
We now describe how individual phase models can be combined to obtain the three-phase model of BLDC motors. The total torque is combined from the three phases operating in parallel:
\begin{equation} \label{eq:total_torque}
    \bm \tau = (K^\phi_{t,A}(\theta) I_A  +  K^\phi_{t,B}(\theta) I_B  +K^\phi_{t,C}(\theta) I_C) \bm j 
\end{equation}
which is the summation of each per-phase torque, where the phase torque constants are identical and 120$^{\circ}$ out of phase: 
\begin{align} \label{eq:kt_out_of_phase}
      K^\phi_{t,A}(\theta)= &-\bar{K^\phi_t}\sin(\theta) \\
     K^\phi_{t,B}(\theta) = &-\bar{K^\phi_t}\sin{(\theta-\frac{2\pi}{3})} \\
     K^\phi_{t,C}(\theta) = &-\bar{K^\phi_t}\sin{(\theta+\frac{2\pi}{3})}
\end{align}

Similarly, the total back-EMF of the BLDC motor becomes:
\begin{equation}
   \boxed{ V_e  =  K^\phi_{b,A} (\theta)  \frac{d\theta_r}{dt} + K^\phi_{b,B} (\theta)  \frac{d\theta_r}{dt} + K^\phi_{b,C} (\theta)   \frac{d\theta_r}{dt}}
\end{equation}
We can apply KVL on each phase quantity, which can be compactly expressed in matrix form (Appendix \ref{sec:appendix_motorphysics}).

\subsection{Electric commutation for BLDC motors}
\begin{figure}[t]
\centering 
 \includegraphics[width=1\columnwidth]{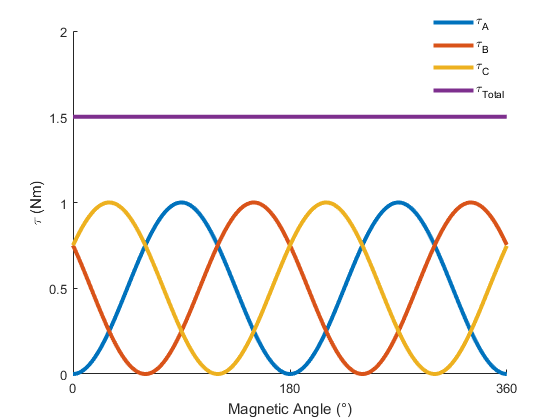}
\caption{A demonstration of torque production of a BLDC motor (\ref{eq:total_torque}). The sum of three phase-torques with constant amplitude results in total torque which is a DC value. Note that each winding produces a torque with a sine squared profile, which results from the product of the sinusoidal torque constant and the sinusoidal phase currents.  }
\label{figure:torque32}
\end{figure}

To produce constant torque, BLDC motors require their windings to be sequentially energized at the appropriate time, a process known as  electric commutation. This process is performed by a microcontroller known as a brushless motor drive.
To maximize the torque production and minimize torque ripple, the driver should provide a current waveform that matches the back-EMF profile \cite{mevey2009sensorless}. That is, for three-phase BLDC motors with sinusoidal back-EMF profiles, the optimal currents to maximize smooth torque production should be in phase with the back-EMF and sinusoidal:
\begin{align} \label{eq:phase_currents}
    I_A = -\bar{I^\phi}\sin{\theta}\\ 
    I_B = -\bar{I^\phi}\sin{(\theta-\frac{2\pi}{3})}\\ 
    I_C = -\bar{I^\phi}\sin{(\theta+\frac{2\pi}{3})}
\end{align}
where $\bar{I^\phi}$ is the amplitude of the phase current and currents are 120$^\circ$ out of phase, identical to the phase of torque constants (\ref{eq:kt_out_of_phase}). For motors with a trapezoidal back-EMF profile, phase currents that match the profile as closely as possible (\textit{i.e.} trapezoidal commutation) should be applied to maximize efficient torque production \cite{cros2002synthesis}. By applying these sinusoidal currents to the total torque (\ref{eq:total_torque}, Fig. \ref{figure:torque32}), the torque becomes:
\begin{equation} 
\tau = \bar{K_t^\phi} \bar{I^\phi} \big[\sin^2(\theta) + \sin^2(\theta+\frac{2}{3}\pi) + \sin^2(\theta-\frac{2}{3}\pi) \big] \bm j
\end{equation}
\begin{equation} \label{eq:torqe_simplified}
   \boxed{ \tau= \frac{3}{2}\bar{K_t^\phi} \bar{I^\phi} \bm j}
\end{equation}
by using trigonometry, shown below:
\begin{equation} \label{trigo}
    \sin^2(\theta) + \sin^2(\theta+\frac{2}{3}\pi) + \sin^2(\theta-\frac{2}{3}\pi) = \frac{3}{2}
\end{equation}

This simplification consists of only DC quantities (\textit{i.e.} torque scales linearly with amplitude of the phase current) and allows us to represent BLDC motors using the brushed motor model. Some literature concludes the conversion to a single phase motor analogue with this equation (\ref{eq:torqe_simplified}) by defining the brushed motor equivalent torque constant as $K_t=\frac{3}{2} \bar{K^{\phi}_t}$ and the amplitude of the phase current ($\bar{I^{\phi}}$) as the brushed motor equivalent current \cite{mevey2009sensorless}. These different definitions of torque constants are valid for calculating torque as long as the appropriate currents are used. Consequently, in this work, we provide a version of the single phase motor representation of a BLDC motor which not only satisfies (\ref{eq:torqe_simplified}), but also enables accurate modeling of other factors (\textit{e.g.} resistive power loss, back-EMF). Methods for appropriate conversions are detailed in the following sections. 

\subsection{d-q transformation}
The power-invariant d-q transformation converts brushless motor quantities to a single phase ``brushed" motor equivalent representation while preserving power of the system\footnote{Sometimes, a different form of the d-q transformation is used which scales the currents such that the q-axis current is equal to the peak of a single phase current \cite{orourke2019dq}. Since this amplitude-invariant form is a non-unitary operation, it renders power calculations inaccurate and is therefore avoided in this analysis.}. The power invariant d-q transformation is a unitary transformation that converts three phase axes in the stator frame of the motor to two axes fixed to a rotating coordinate frame that rotates with the rotor's magnetic angle (\textit{i.e.} d-q axis). Conversion to the rotating reference frame of the rotor transforms sinusoidal functions of angle, such as current and voltage, into effective DC values, forming an equivalent single phase brushed motor analogue for BLDC motors. The d-q transformation is composed of two transformations: 

\begin{equation} \label{eq:clarke}
\text{Clarke ($\bm C$): }
\sqrt{\frac{2}{3}}
\begin{bmatrix}
1 &- \frac{1}{2} &- \frac{1}{2} \\
0 &\frac{\sqrt{3}}{2} &-\frac{\sqrt{3}}{2} \\
\end{bmatrix} 
\end{equation}
\begin{equation} \label{eq:park}
\text{Park ($\bm P$):}
\begin{bmatrix}
\cos(\theta) &\sin(\theta)\\
-\sin(\theta) &\cos(\theta)
\end{bmatrix}
\end{equation}
where \textbf{C} and \textbf{P} are Clarke and Park transformations, respectively, and the matrix product of these transformations becomes the d-q transformation:
\begin{equation} \label{eq:dq_trans}
\boxed{\text{ $\bm P\cdot \bm C$= }
\sqrt{\frac{2}{3}}
\begin{bmatrix}
\cos(\theta) &\cos(\theta-\frac{2}{3}\pi) &\cos(\theta + \frac{2}{3}\pi) \\
-\sin(\theta) &-\sin(\theta-\frac{2}{3}\pi) &-\sin(\theta + \frac{2}{3}\pi)
\end{bmatrix}}
\end{equation}
By applying the d-q transformation to phase currents or voltages, the converted quadrature quantities are described as follows:
\begin{equation}
\begin{gathered}
\begin{bmatrix}
I^d \\
I^q \\
\end{bmatrix}
= \bm P\cdot \bm C \cdot \bm I^\phi = \\
\sqrt{\frac{2}{3}}
\begin{bmatrix}
\cos(\theta) &\cos(\theta-\frac{2}{3}\pi) &\cos(\theta + \frac{2}{3}\pi) \\
-\sin(\theta) &-\sin(\theta-\frac{2}{3}\pi) &-\sin(\theta + \frac{2}{3}\pi)
\end{bmatrix}
\cdot
\begin{bmatrix}
I^\phi_A\\
I^\phi_B\\
I^\phi_C
\end{bmatrix}
\end{gathered}
\end{equation}
\[
= 
\sqrt{\frac{2}{3}}
\begin{bmatrix}
0 \\
\frac{3}{2} \bar{I^\phi} 
\end{bmatrix}
\]

Thus, the q and d-axis currents are 
\begin{equation} \label{eq:quad_current_conv}
    \boxed{I^q = \sqrt{\frac{3}{2}} \bar{I}^\phi}
\end{equation}
\begin{equation} \label{eq:quad_current_conv}
    I^d = 0
\end{equation}
and the BLDC motor torque can be expressed as: 
\begin{equation} \label{eq:kt_q}
\bm \tau_q = \bar{K}^\phi_{t}\sqrt{\frac{3}{2}}  \bar{I}^\phi\sqrt{\frac{3}{2}} \bm j = K^q_t  I^q \bm j
\end{equation}
where the q-axis torque constant is defined as:
\begin{equation} \label{eq:quad_kt_conv}
   \boxed{ \bar{K}^\phi_t\sqrt{\frac{3}{2}}= K^q_t}
\end{equation}
and superscript $q$ denotes q-axis quantities of according constants, while the d-axis current is zero. Similarly, 
we can define the back-EMF constant using the same conversion: 
\begin{equation}
K_b^q = \sqrt{\frac{3}{2}} \cdot K_b^\phi
\end{equation}
where the phase back-EMF constant and phase torque constant are identical ($K_t^\phi=K_b^\phi$). This relationship shows that there is factor of $\sqrt{\frac{3}{2}}$ difference among quadrature and phase quantities. Therefore, by applying Newton's second law, we can obtain:
\begin{equation}  \label{eq:brushed_mech}
 \boxed{J\frac{d^2\theta_r}{d^2t} = K^q_t  I^q -  b \frac{d\theta_r}{dt} - \tau_L}
\end{equation}
Similarly, by applying the d-q transformation, the equivalent electrical circuit of a BLDC motor becomes:
\begin{equation} \label{eq:quad_electric}
   \boxed{ V^q = R^{\phi} I^q + K^q_b\frac{d\theta_r}{dt} + L^e \frac{dI_q}{dt}}
\end{equation}
where, $L^e$ is an effective inductance (\textit{i.e.} stator inductance is invariant of rotor position \cite{koteich2013real}).  For full derivation, please refer to Appendix \ref{section:appedix_a}. Since neither self inductance nor mutual inductance may be accessible, the effective inductance can be obtained using the terminal inductance as follows:
\begin{empheq}[box=\fbox]{align} 
   \text{Wye:}&\;\;\;\;\;\;\;\;L^{e}=\frac{3}{2} L^{ll}\label{eq:inductance_wye}\\
    \text{Delta:}&\;\;\;\;\;\;\;\;L^{e}=\frac{1}{2} L^{ll}\label{eq:inductance_delta}
\end{empheq}
where, $L^{ll}$ represents the line-to-line or terminal inductance. This assumes the connection scheme of phases as Fig. 1 of \cite{musak2013novel}. In this work, we assume identical self and mutual inductances across different phases and their combinations. By applying the d-q transformation, the effective phase inductance is equal to the q-axis inductance (Appendix \ref{section:appedix_a}):
\begin{equation}
L^e= L^{q}
\end{equation}
The d-q axis representation of a BLDC motor demonstrates that the motor can be fully transformed to the single phase ``brushed" motor representation that is helpful for analysis and control.

\begin{figure}[t]
\centering 
 \includegraphics[width=1.0\columnwidth]{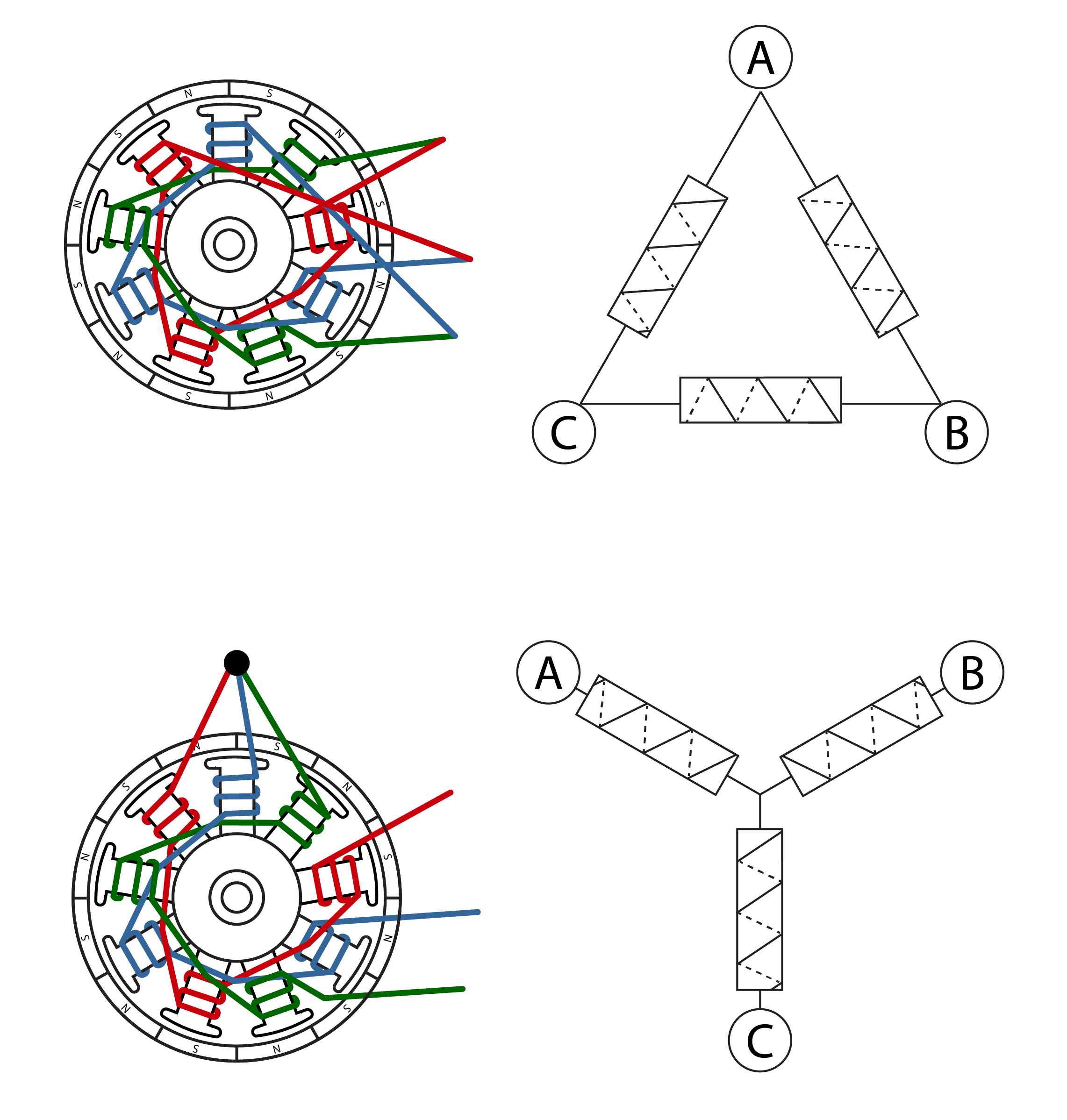}
\caption{Depictions of delta (top) and wye (bottom) winding configurations.}
\label{fig:winding_config}
\end{figure}

\subsection{BLDC motor winding configuration} \label{sec:winding_config}
To fully understand the relationships that govern BLDC motors and their characteristics, we must first understand the motor's winding type. There are two types of common BLDC windings: wye (\textit{i.e.} star) and delta (Fig. \ref{fig:winding_config}). The winding type of BLDC motors is determined by the manufacturer and can be chosen for a variety of factors. Delta wound motors are generally used when greater instantaneous current or torque is required. If equal line-to-line voltages are applied to both types of windings, the entire voltage is applied across each phase for a delta wound motor ($V^{ll} = V^{\phi})$, whereas the line-to-line voltage is divided between the phases for a wye wound motor ($V^{ll}/\sqrt{3} = V^{\phi}$). In addition, wye wound motors typically have greater efficiency in the lower velocity regions, while delta wound motors are typically used when greater efficiencies are desired in higher velocity regions \cite{lee2014phase}. 
Finally, specific winding types may have benefits in certain applications (\textit{e.g.} induction motors) \cite{mevey2009sensorless}.  

The winding type dictates how terminal (\textit{i.e.} line-to-line) quantities are related to phase quantities. Lines are defined as the three conductors that branch from the windings (A, B, and C), and line-to-line voltage is defined as the voltages referenced between any two lines. The relationship between the phase quantities and line-to-line quantities for each winding configuration are as follows:
\begin{empheq}[box=\fbox]{align}
\text{Wye:}\;\;\;\;\;\;\;\;\;\; \label{eq:wye_voltage1} \bar{V}^{ll} = &\sqrt{3}\bar{V}^{\phi}\;\; \\ \bar{I}^l= & \bar{I}^{\phi} \label{eq:wye_voltage2} \\
 \text{Delta:}\;\;\;\;\;\;\;\;\; \bar{V}^{ll} = &\bar{V}^{\phi}\;\; \label{eq:delta_voltage1}\\     \bar{I}^l = & \sqrt{3}\bar{I}^{\phi} \label{eq:delta_voltage2}
\end{empheq}
Note that currents are referenced within a line, and voltages are measured between lines (\textit{i.e.} there is no  line-to-line current). Depending on the winding type, appropriate conversion factors should be applied to accurately convert between reference frames (\textit{e.g.} d-q transformation). Thus, knowing the winding type is important for estimating the motor properties, which ultimately enables accurate modeling, analyses, and commutation\footnote{Brushless motor drives that perform field oriented control and report q-axis current and voltage often assume one motor winding type (\textit{e.g.} wye) to perform their calculations. If the motor being used is delta wound, appropriate conversions factors will be needed to ensure accurate estimation of expected motor torque.} for robotics applications.   
\subsection{Resistive power loss} \label{sec:power_loss}
The heat produced from a motor's windings are among the chief limiting factors in a motor's performance. Thus, to appropriately size a motor for a specific task, the resistive power loss and temperature rise of the windings is often investigated\footnote{ The analysis developed in this paper has neglected other sources of losses such as core losses due to hysteresis and eddy currents, bearing friction, air resistances \textit{etc.} \cite{hendershot2010design}. We chose to focus on Joule heating because it is the dominant factor in temperature rise, and it can be challenging to obtain information for other sources of losses from manufacturer datasheet.}. We can calculate the total resistive heating loss by summing individual losses of each phase:
\begin{align} \label{eq:power_phase}
P = & ({I^{\phi}_A}^2 + {I^{\phi}_B}^2 + {I^{\phi}_C}^2) R^\phi \\
  = & \frac{3}{2} \bar{I^\phi}^2  R^\phi 
\end{align}
where each phase currents are provided as (\ref{eq:phase_currents}). Similar to the total torque, the total power loss becomes a DC value. The power loss can also be represented in the q-axis frame:
\begin{equation} \label{eq:power_quad}
\boxed{P = {I^{q}}^2 R^\phi}
\end{equation}
where it reduces to the single phase brushed motor equivalent expression by using the conversion factors (\ref{eq:quad_current_conv}, \ref{eq:quad_kt_conv}). Note that resistance used in the q-axis power loss formulation is the resistance of a single phase.
\newcolumntype{C}{>{\centering\arraybackslash}X}
\begin{table}[t] 
\caption {Brushless Motor Specifications} \label{tab:spec}
\centering
\begin{threeparttable}
\begin{tabular}{L{3cm}L{4cm}} 
\hline
Characteristics	& Definition \& Comments \\
\hline
             &   \\
Terminal Resistance ($\Omega$)        & Represents electrical resistance across two motor leads  (line-to-line or phase-to-phase resistance). Note that to estimate the phase resistance (neutral-to-line), one should know the winding configuration of the motor, which is typically not provided by the motor manufacturers. Appendix \ref{sec:appendix_identify_winding_config} details how to determine the winding type.     \\\\
Torque Constant, $K_t$ (Nm/A)             & Ratio of the torque output to the provided current. For many BLDC motors, the definition of what current is used to determine the torque constant is not provided (\textit{e.g.} whether the current used is the phase current or line current). \\\\
Back-EMF Constant, $K_b$ (V/(rad/s)) & Induced voltage created by rotating the motor's rotor. Typically provided in the line-to-line reference frame, however; often the representation information is omitted in motor datasheets.  \\
\hline
\end{tabular}
\end{threeparttable}
\end{table}  

\section{Implications in BLDC Analyses}

\begin{table}[h]
    \centering
    \caption{Common representations of current and voltage in brushless motors}
    \begin{tabular}{L{3cm}L{4cm}}
    \hline
    Notation&Description \\
    \hline
    $I^q,V^q$ & q-axis current / voltage, equivalent to RMS current / voltage of all three phases$^\dagger$  \\\\
    $I^{\phi},V^{\phi}$ & Peak current/voltage of one phase, equivalent to q-axis current/voltage using the power-varying transform  \\\\
    $I^{l},V^{ll}$ & Line current and line-to-line voltage, measured within or across two of the motor leads, respectively. Conversion factors vary depending on winding type. \\\\
    $I^{RMS},V^{RMS}$ & RMS current/voltage for one phase \\
    \hline \\
    \end{tabular}
    \label{tab:signals}
    \begin{tablenotes}
    \item $^\dagger$ $
\sqrt{\frac{\int_{0}^{T} (I^{\phi^2}_A + I^{\phi^2}_B + I^{\phi^2}_C)}{T}dt} 
=\sqrt{\frac{3}{2}}\bar{I^{\phi}} = I^q $
    \end{tablenotes}
\end{table}

\subsection{BLDC datasheet variation}
To optimally select BLDC motors for robotics applications, \textit{a priori} modeling analyses are conducted to properly specify the motor for the application; this is especially important when motor mass is a key driver of system performance. The specific parameters needed to conduct these analyses are often obtained from manufacturer datasheets.
Inconsistencies and sparse or ambiguous information provided in these datasheets can lead to erroneous results (Table \ref{tab:spec}). That is, manufacturers sometimes report motor parameters with respect to different voltage and current references. Indeed, these parameters are often measured based on their simplicity or convenience, as opposed to what would most readily be used in modeling analyses. Several common reference frames are described in Table \ref{tab:signals}. In the proceeding sections, we describe common errors that arise from improperly using motor parameters in BLDC modeling analyses. We provide conversions from common values of voltage and current, and encourage motor manufacturers to specify the details of the provided parameters with emphasis on reporting with respect to the q-axis representation.

\subsubsection{Thermal loss, resistance, and inductance}
Thermal loss is a key driver of electromechanical performance; to minimize motor mass, the windings are expected to rise in temperature, without exceeding the maximum permissible temperature. To correctly model the resistive heat loss, the heat produced by all three windings is considered (\ref{eq:power_phase}). Equivalently, the thermal loss can be calculated from a single phase ``brushed" motor analogue (\ref{eq:power_quad}) using the q-axis current and phase resistance. However, the resistance and current values used in these equations must be paired appropriately; when modeling the resistive heat loss using the q-axis current, the required winding resistance is the phase resistance. This is different from the motor's terminal resistance, which can be reported on BLDC motor datasheets. To convert from terminal resistance to phase resistance, the winding type must be considered (Appendix \ref{sec:appendix_identify_winding_config}). However, this conversion can be challenging, since many BLDC motor datasheets do not include the motor winding type, and thus obscure the ability to calculate phase resistance from terminal resistance. As an example of how resistive heat loss can be erroneously calculated by pairing the incorrect resistance and current, if terminal / line-to-line resistance is used directly instead of the phase resistance, the predicted power loss will be twice the magnitude for wye-wound motors and 33\% less for delta-wound motors (Appendix \ref{sec:appendix_power_inaccurate}).

To address this issue, terminal resistance can be scaled to the appropriate phase resistance by (\ref{eq:resistance_wye}) and (\ref{eq:resistance_delta}), and used to calculate total resistive power loss (\ref{eq:power_quad}):

\begin{empheq}[box=\fbox]{align}
    \text{Wye:}&\;\;\;\;\;\;\;\;R^{\phi}=\frac{1}{2} R^{ll}\label{eq:resistance_wye}\\
    \text{Delta:}&\;\;\;\;\;\;\;\;R^{\phi}=\frac{3}{2} R^{ll}\label{eq:resistance_delta}
\end{empheq}

Following the determination of the appropriate resistance values to use in power loss calculations, the expected rise in temperature can be obtained in simulation using the thermal resistances and capacitances for the coupled winding-housing system dynamics\cite{lee2019empirical}. 

\subsubsection{Torque} \label{sec:torque}
The calculation of expected motor torque is a critical aspect of the design process. There are two common errors that result from inaccurate usage of motor parameters; firstly, some recently developed motors from the drone industry provide the velocity constant ($K_v$) rather than torque constant, and this is sometimes erroneously converted directly to the motor's torque constant. Secondly, the current used in the formulation of the motor's torque constant reported in the datasheet may be ambiguous, and can cause torque estimation error if the current used to determine the torque constant does not agree with the current used to predict torque in modeling analyses. To address these challenges, appropriate conversion factors and explanations are provided in the paragraphs below.\par
To calculate expected motor torque, a single phase ``brushed" motor analogue is typically used, in which the motor's torque constant is multiplied by the effective winding current. In some cases, motor manufacturers provide the voltage constant as a proxy for the torque constant, and leave the torque constant determination to the designer since these values are equivalent with proper commutation (\textit{i.e.} the manufacturer only provides the reciprocal of the motor's $K_b$, known as $K_v$). This may lead to a common error---considering $K_b$ equal to $K_t$---which is not necessarily true for brushless motors \cite{hendershot2010design}, depending on the winding reference frame. For example, a common motor used in some robotic applications is the T-motor KV100 (U8), where only the velocity constant $K_v$ ($\frac{1}{K_b}$) is provided as 10.47 rad/V$\cdot$s (100 RPM/V). However, the reciprocal of this value actually (nearly) describes the amplitude of the line-to-line sinusoidal back-EMF per unit angular velocity (\textit{i.e.} $\bar{K_b^{ll}}$), which, because the motor is delta wound, represents the back-EMF amplitude for a single phase as a function of rotor velocity (\ref{eq:delta_voltage1}). However, in practice, the amplitude of the line-to-line back-EMF is $\sim$5\% to 10\% greater than what would be expected using the manufacturer reported $K_b^{ll}$ (potentially converted from $K_v^{ll}$). To obtain the q-axis torque constant for an equivalent single phase ``brushed" motor, the amplitude-based $\bar{K_b^{ll}}$ is multiplied by $\sqrt{\frac{3}{2}}$, which can then be used with the q-axis current. Refer to Appendix \ref{section:appendix_torque_inaccurate} for full derivation. Generalized for both winding types, the conversions are:
\begin{empheq}[box=\fbox]{align}
\text{Wye:}&\;\;\;\;\;\;\;\; K^q_t = \frac{1}{\sqrt{2}}\bar{K_b^{ll}} \label{eq:Kt_conversion_wye}\\
\text{Delta:}&\;\;\;\;\;\;\;\ K^q_t = \sqrt{\frac{3}{2}}\bar{K_b^{ll}}
\label{eq:Kt_conversion_delta}
\end{empheq}

The BLDC motor torque can be also calculated using the velocity constant, which is agnostic to winding type:
\begin{equation} \label{eq:kb_ll_torque}
   \boxed{\tau = \frac{\sqrt{3}}{2}  \bar{K_b^{ll}}  \bar{I^{l}}}
\end{equation}
which can be derived from equations \ref{eq:q_axis_torque}, \ref{eq:Kt_conversion_wye}, \ref{eq:Kt_conversion_delta}, \ref{eq:current_wye} and \ref{eq:current_delta}.

Modeling expected motor torque can be further hindered by unclear representations of torque constants from manufacturer datasheets. Informal communications revealed several different current conventions used when reporting torque constants, which were rarely explained on their datasheets or product literature. While our understanding of the currents used by these manufacturers is reported in Table \ref{tab:manufacturers}, it should be noted that it was challenging to obtain the exact definitions of the currents used in torque constants (which illustrates the issue) though we are grateful to the manufacturers who worked with us. Thus, care must be used when interpreting torque constants, and we recommend conversion of torque constants to the q-axis representation for accurate and convenient modeling analysis. The following equations can be used to convert torque constants provided with different current references to the equivalent torque constant using the q-axis winding reference frame:
\begin{table}[h]
   \renewcommand{\arraystretch}{1.5}
    \centering
    \caption{Different manufacturers' representation of torque constants}
    \begin{tabular}{L{4cm}L{4cm}}
    \hline
    Manufacturer&Current in $K_t$\\
    \hline
    Maxon & $\bar{I^{\phi}}$ \\
    T-Motor & $I^{bus}$ \\
    TQ Drive & $\bar{I^{\phi}}$ \\
    Micromo $^\dagger$  & $I^{RMS}      $ \\
    Kollmorgen & $I^q$ \\
    Parker & $I^{RMS}$ \\
    \hline 
    \end{tabular}
    {\begin{flushleft}Note that the above description is based on informal conversations with engineers from each manufacture, not from official product documentation (it was unlisted). In addition, the table only applies to FOC/Sine commutation. \end{flushleft}}
    {\begin{flushleft} $^\dagger$ It is unclear whether they use RMS of three phases or a single phase, which should be further investigated. \end{flushleft}}
    \label{tab:manufacturers}
\end{table}
\begin{empheq}[box=\fbox]{align}
\text{Wye:}&\;\;\;\;\;\;\;K^q_{t} = \sqrt{\frac{3}{2}}\bar{K^\phi_{t}}\label{eq:conversion_kt_wye}\\
\text{Delta:}&\;\;\;\;\;\;\;K^q_{t} = \sqrt{\frac{3}{2}}\bar{K^\phi_{t}} \label{eq:conversion_kt_delta}
\end{empheq}

For convenience, we also provide the equations for q-axis current obtained using the different winding reference frames:
\begin{empheq}[box=\fbox]{align}
\text{Wye:}&\;\;\;\;\;\;\;I^q = \sqrt{\frac{3}{2}}\bar{I}^{\phi} = \sqrt{\frac{3}{2}}\bar{I}^l \label{eq:current_wye}\\
\text{Delta:}&\;\;\;\;\;\;\;I^q = \sqrt{\frac{3}{2}}\bar{I}^{\phi} =  \frac{1}{\sqrt{2}}\bar{I}^l \label{eq:current_delta}
\end{empheq}

Thus, using the appropriately paired q-axis current and torque constant, the expected torque can be modeled by:
\begin{equation}
    \boxed{\tau = K_t^qI^q \label{eq:q_axis_torque}}
\end{equation}

\subsubsection{Voltage} \label{sec:voltage_implication}
An important factor in the design process is choosing the required bus (\textit{i.e.} battery or power supply) voltage. The bus voltage governs the maximum velocity at which a motor can operate, assuming the motor is operating within other speed restrictions (\textit{e.g.} bearing maximum angular velocity). In a brushed ``single phase" motor, the calculation of maximum no-load velocity is trivial, simply obtained using the bus voltage divided by the back-EMF constant. However, when analyzing brushless motors, obtaining the maximum velocity requires a more detailed analysis. To determine the maximum no-load velocity, we need to consider the maximum amplitude of the motor's line-to-line voltage, governed by the motor's winding type. The amplitude of the line-to-line voltage may exceed the q-axis voltage, and thus designing based on the q-axis voltage may result in unexpected limitations to motor velocity\footnote{Some BLDC drives will modify the commutation routine when $\bar V^{ll}$ is saturated at $V^{bus}$--this enables the motor to achieve a greater allowable velocity for a fixed $V^{bus}$. When this occurs, the voltage and current profiles can be become non-sinusoidal. This alteration generally causes a non-zero d-axis current that effectively modifies the motor's velocity constant. This approach can be used to increase or decrease the motor's velocity constant for a corresponding loss in efficiency but is outside the scope of this tutorial.}. The amplitude of the motor's line-to-line voltage can be obtained using (\ref{eq:voltage_wye}) and (\ref{eq:voltage_delta}):
\begin{empheq}[box=\fbox]{align} 
\text{Wye:}&\;\;\;\;\;\;\;\bar{V}^{ll} = \sqrt{2}V^{q} = \sqrt{2}K^q_b\frac{d\theta_r}{dt}\label{eq:voltage_wye}\\
\text{Delta:}&\;\;\;\;\;\;\;\bar{V}^{ll} = \sqrt{\frac{2}{3}}V^{q} = \sqrt{\frac{2}{3}}K^q_b\frac{d\theta_r}{dt}\label{eq:voltage_delta}
\end{empheq}
where $K^q_b$ is the q-axis back-EMF constant (equivalent in magnitude to the q-axis torque constant) and $V^{q}$ is obtained using the back-EMF constant and angular velocity. Solving these equations for velocity and substituting $V^{bus}$ for $\bar{V}^{ll}$ provides an estimate for the maximum angular velocity for a given bus voltage under no load conditions:
\begin{empheq}[box=\fbox]{align}
\text{Wye:}&\;\;\;\;\;\;\;\;\frac{d\theta_r}{dt}_{max} = \sqrt{\frac{1}{2}}\frac{V^{bus}}{K^q_b}\\
\text{Delta:}&\;\;\;\;\;\;\;\;\frac{d\theta_r}{dt}_{max} = \sqrt{\frac{3}{2}}\frac{V^{bus}}{K^q_b}
\end{empheq}
In addition, there are other methods to calculate maximum velocity (\textit{e.g.} using phase quantities, including $\bar{K^\phi_b}$), however, we recommend using the q-axis representation for consistencies with other analyses. 

\subsection{Standardizing Motor Analysis}
Accurate modeling of the torque, velocity, current, and voltage needed for robotic applications is an important step in the design process. To ensure motor manufacturers provide the necessary information needed to perform these analyses, we suggest a standard set of information should be provided. Ideally, motor manufacturers would report information with respect to the q-axis reference frame, however, if this is impractical, we recommend the following. Datasheets could include line-to-line (terminal) resistance and inductance, in addition to the winding type, which enables conversion to the q-axis reference frame. The torque constant can be provided with respect to any winding reference frame, but the current used in the quantification of torque constant must be clearly described. In addition, while some manufacturers provide information on thermal dynamics (\textit{e.g.} winding / housing thermal resistances and time constants), many omit this information, leaving this characterization to the community. Overall, each parameter should clearly indicate how it relates to a specific winding reference frame, and, if winding type is provided, conversions can be performed with ease. 

\section{Conclusion}
In this tutorial, we present the underlying mathematical modeling of brushed and brushless DC motors, as well as electric commutation and winding configurations of BLDC motors. We described potential sources of error in BLDC motor modeling, which often stem from inconsistencies and misinterpretation of manufacturer datasheets. To address these errors, we provide explanations and conversions to the direct-quadrature reference frame, which facilitates a convenient DC representation while conserving key motor properties (\textit{e.g.} resistive power loss). The intent of this tutorial is two fold: 1. to guide engineers and robot designers who utilize BLDC motors in their application to more accurately model and select a motor that is optimal for their application; and 2. to highlight the need for greater standardization and details to be provided in manufacturer datasheets (\textit{e.g.} unit clarity, winding configuration, \textit{etc.}). We hope that this tutorial can serve as a benchmark for standardizing BLDC motor modeling analyses while steering engineers and designers to more accurate and high performing system designs. 


\appendices
\section{BLDC motor modeling cheat sheet} \label{sec:summary}
This appendix provides a condensed set of equations that are intended to be more easily referenced when analyzing BLDC motors in the design process. As a reminder, these equations are intended for brushless motors with a sinusoidal back-EMF profile. The cheat sheet includes governing equations of brushed and BLDC motors, and conversions for representing BLDC motors as a single phase ``brushed" motor analogue, which includes a single current, voltage resistance, inductance, and torque / velocity constants, which are more convenient to analyze during modeling and design specification. We advocate for the DC representation (\textit{i.e.} q-axis) of BLDC motors, which would provide convenient and accurate analysis while conserving critical properties, such as resistive power loss and torque production. 
For full derivation of the equations, please refer to the main sections of the paper, where the equations included in this cheat sheet are denoted with boxes. Equation numbers follow the numbering in the main sections.    

\subsection{Prerequisites}
To convert to the q-axis DC representation of a selected BLDC motor, the designer should identify at least one frame of reference (q-axis, phase, line) for each electrical quantity. These quantities include $R, K_t, K_v$, which typically can be obtained from manufacturer datasheets; either (desired) current / voltage or torque / velocity in a specified reference frame; as well as the motor's winding type (wye or delta). In addition, we provide a procedure for identifying the winding type of the motor in case they are not included in the manufacturers' specification (Appendix \ref{sec:appendix_identify_winding_config}).

\resttable

\subsection{Governing motor equations}

\subsubsection{Brushed DC motor}
\begin{equation} \tag{\ref{eq:brushed_mech}}
     J\frac{d^2\theta_r}{d^2t} = K_t I^a -  b \frac{d\theta_r}{dt} - \tau_L
\end{equation}
\begin{equation} \tag{\ref{eq:brushed_elec}}
V = RI^a + K_b \frac{d\theta_r}{dt} + L\frac{dI^a}{dt} 
\end{equation}
\subsubsection{Brushless DC motor}

\begin{equation} \label{eq:quad_mech}
    J\frac{d^2\theta_r}{d^2t} = K_t^q I^q -  b \frac{d\theta_r}{dt}-\tau_L
\end{equation}
\begin{equation} \tag{\ref{eq:quad_electric}}
    V^q = R^{\phi}\cdot I^q + K^q_b\frac{d\theta_r}{dt} + L^e \frac{dI_q}{dt}
\end{equation}
\makeatletter
\newcommand*\bigcdot{\mathpalette\bigcdot@{.5}}
\newcommand*\bigcdot@[2]{\mathbin{\vcenter{\hbox{\scalebox{#2}{$\m@th#1\bullet$}}}}}
\makeatother

\begin{equation} \tag{\ref{eq:dq_trans}}
\begin{gathered}
\begin{bmatrix}
\bigcdot^d \\
\bigcdot^q \\
\end{bmatrix}
= \bm P \bm C  \begin{bmatrix}
\bigcdot^\phi_A\\
\bigcdot^\phi_B\\
\bigcdot^\phi_C
\end{bmatrix} \\
\bm P \bm C = \\ 
\sqrt{\frac{2}{3}}
\begin{bmatrix}
\cos(\theta) &\cos(\theta-\frac{2}{3}\pi) &\cos(\theta + \frac{2}{3}\pi) \\
-\sin(\theta) &-\sin(\theta-\frac{2}{3}\pi) &-\sin(\theta + \frac{2}{3}\pi)
\end{bmatrix}
\end{gathered}
\end{equation}
where $\bigcdot$ denotes an arbitrary electrical quantity (\textit{e.g.} current or voltage). 

\subsection{Conversion to q-axis quantities} \label{sec:q-conversion}

\begin{align}
\text{Wye:}\;\;\;\;\;\;\;\;\;\; \bar{V}^{ll} = &\sqrt{3}\bar{V}^{\phi}\;\; \tag{\ref{eq:wye_voltage1}}\\ \bar{I}^l= & \bar{I}^{\phi} \tag{\ref{eq:wye_voltage2}}\\
 \text{Delta:}\;\;\;\;\;\;\;\;\; \bar{V}^{ll} = &\bar{V}^{\phi}\;\; \tag{\ref{eq:delta_voltage1}}\\     \bar{I}^l = & \sqrt{3}\bar{I}^{\phi} \tag{\ref{eq:delta_voltage2}}
\end{align}

\subsubsection{Resistance, resistive power loss, \& inductance}

\begin{align}
    \text{Wye:}&\;\;\;\;\;\;\;\;R^{\phi}=\frac{1}{2} R^{ll}\tag{\ref{eq:resistance_wye}}\\
    \text{Delta:}&\;\;\;\;\;\;\;\;R^{\phi}=\frac{3}{2} R^{ll}\tag{\ref{eq:resistance_delta}}
\end{align}

\begin{equation} \tag{\ref{eq:power_quad}}
P = {I^{q}}^2 R^\phi
\end{equation}

\begin{align}
   \text{Wye:}&\;\;\;\;\;\;\;\;L^{q}=\frac{3}{2} L^{ll}\tag{\ref{eq:inductance_wye}}\\
    \text{Delta:}&\;\;\;\;\;\;\;\;L^{q}=\frac{1}{2} L^{ll}\tag{\ref{eq:inductance_delta}}
\end{align}
\subsubsection{Current \& torque}

\begin{table}[h]
    \renewcommand{\arraystretch}{1.8}
    \centering
    \caption{Conversion to phase / q-axis from line / terminal quantities}
    \begin{tabular}{L{1cm}M{2.5cm}M{2.5cm}}
    \hline
    Parameter & \multicolumn{2}{c}{Equivalencies}   \\
            &      Wye &      Delta  \\
    \hline 
    $\bar{V}^{\phi}$ & $\frac{1}{\sqrt{3}} \bar{V}^{ll}$ & $\bar{V}^{ll}$\\
    $\bar{I}^\phi$ & $\bar{I}^{l}$ & $\frac{1}{\sqrt{3}}\bar{I}^{l}$ \\
    $R^{\phi}$ & $\frac{1}{2} R^{ll}$  & $\frac{3}{2} R^{ll}$  \\
    $L^{q}$ & $\frac{3}{2} L^{ll}$  & $\frac{1}{2} L^{ll}$  \\
    $V^q$ & $\frac{1}{\sqrt{2}} \bar{V}^{ll}$ & $ \sqrt{\frac{3}{2}} \bar{V}^{ll}$\\
    $I^q$ & $\sqrt{\frac{3}{2}}\bar{I^l}$ &  $\sqrt{\frac{1}{2}}\bar{I^l}$ \\
    $K^{q}_t$ & $\frac{1}{\sqrt{2}}\bar{K_b^{ll}}$ &  $\sqrt{\frac{3}{2}}\bar{K_b^{ll}}$   \\
    $\tau$  & \multicolumn{2}{c}{$K_t^qI^q$   \text{or} $\frac{\sqrt{3}}{2}  \bar{K_b^{ll}} \bar{I^{l}}$ }  \\
    $P$  & \multicolumn{2}{c}{  ${I^{q}}^2 R^{\phi} $}  \\
    \hline 
    \end{tabular}
    {\begin{flushleft} Tabulated description of Section \ref{sec:q-conversion}. Note, if converting from a manufacturer's reported $K_v$ number ($1/K_b^{ll}$), see Section  \ref{sec:torque}. \end{flushleft}}
    \label{tab:q-conversion}
\end{table}

\begin{align}
\text{Wye:}&\;\;\;\;\;\;\;\sqrt{\frac{3}{2}}\bar{I^l} = \sqrt{\frac{3}{2}}\bar{I^\phi} = I^q \tag{\ref{eq:current_wye}}\\
\text{Delta:}&\;\;\;\;\;\;\;\sqrt{\frac{1}{2}}\bar{I^l} = \sqrt{\frac{3}{2}}\bar{I^\phi} = I^q\tag{\ref{eq:current_delta}}
\end{align}

\begin{align}
\text{Wye:}&\;\;\;\;\;\;\;K^q_{t} = \sqrt{\frac{3}{2}}\bar{K^\phi_{t}}\tag{\ref{eq:conversion_kt_wye}}\\
\text{Delta:}&\;\;\;\;\;\;\;K^q_{t} = \sqrt{\frac{3}{2}}\bar{K^\phi_{t}} \tag{\ref{eq:conversion_kt_delta}}
\end{align}
 \begin{align}
\text{Wye:}&\;\;\;\;\;\;\;\; K^q_t = \frac{1}{\sqrt{2}}\bar{K_b^{ll}} \tag{\ref{eq:Kt_conversion_wye}}\\
\text{Delta:}&\;\;\;\;\;\;\;\ K^q_t = \sqrt{\frac{3}{2}}\bar{K_b^{ll}}
\tag{\ref{eq:Kt_conversion_delta}}
\end{align}
 
\begin{equation}
    \tau = K_t^qI^q \tag{\ref{eq:q_axis_torque}}
\end{equation}

\begin{equation} \tag{\ref{eq:kb_ll_torque}}
\tau = \frac{\sqrt{3}}{2}  \bar{K_b^{ll} \bar{I^{l}}}
\end{equation}

\subsubsection{Voltage \& angular velocity}

 \begin{align} 
\text{Wye:}&\;\;\;\;\;\;\;\bar{V}^{ll} = \sqrt{2}V^{q} = \sqrt{2}K^q_b\frac{d\theta_r}{dt} \tag{\ref{eq:voltage_wye}}\\
\text{Delta:}&\;\;\;\;\;\;\;\bar{V}^{ll} = \sqrt{\frac{2}{3}}V^{q} = \sqrt{\frac{2}{3}}K^q_b\frac{d\theta_r}{dt} \tag{\ref{eq:voltage_delta}}
\end{align}

\begin{align}
\text{Wye:}&\;\;\;\;\;\;\;\;\frac{d\theta_r}{dt}_{max} = \sqrt{\frac{1}{2}}\frac{V^{bus}}{K^q_b} \tag{\ref{eq:voltage_wye}}\\
\text{Delta:}&\;\;\;\;\;\;\;\;\frac{d\theta_r}{dt}_{max} = \sqrt{\frac{3}{2}}\frac{V^{bus}}{K^q_b} \tag{\ref{eq:voltage_delta}}
\end{align}

\section{Winding Type identification procedure} \label{sec:appendix_identify_winding_config}
In this tutorial, one of the key attributes needed is the BLDC motor's winding type. Often, this information is not included in motor specifications or datasheets. In this case, we provide a procedure for identifying the winding type as described below.

\subsubsection{Contact the manufacturer}
The simplest method of identifying the winding type of a BLDC motor is to contact the motor manufacturer. Importantly, wye wound motor types may be called ``star" or ``star-serial".
\subsubsection{Visual Inspection}
Another convenient method for identifying the motor's winding type when the windings are visible is to visually inspect them. For wye wound motors, the leads of the three phases are connected to form a neutral point (see Fig. \ref{fig:winding_config}) with the other ends of the phases being the motor's leads. Sometimes, this neutral point can be observed (\textit{i.e.} under shrink wrap). Alternatively, delta wound motors have no neutral points and the combination of each of the phases are wired together to form the ends (see Fig. \ref{fig:winding_config}).

\subsubsection{Thermal Imaging}
\begin{figure}[t]
\centering 
\includegraphics[width=1.0\columnwidth,angle=180]{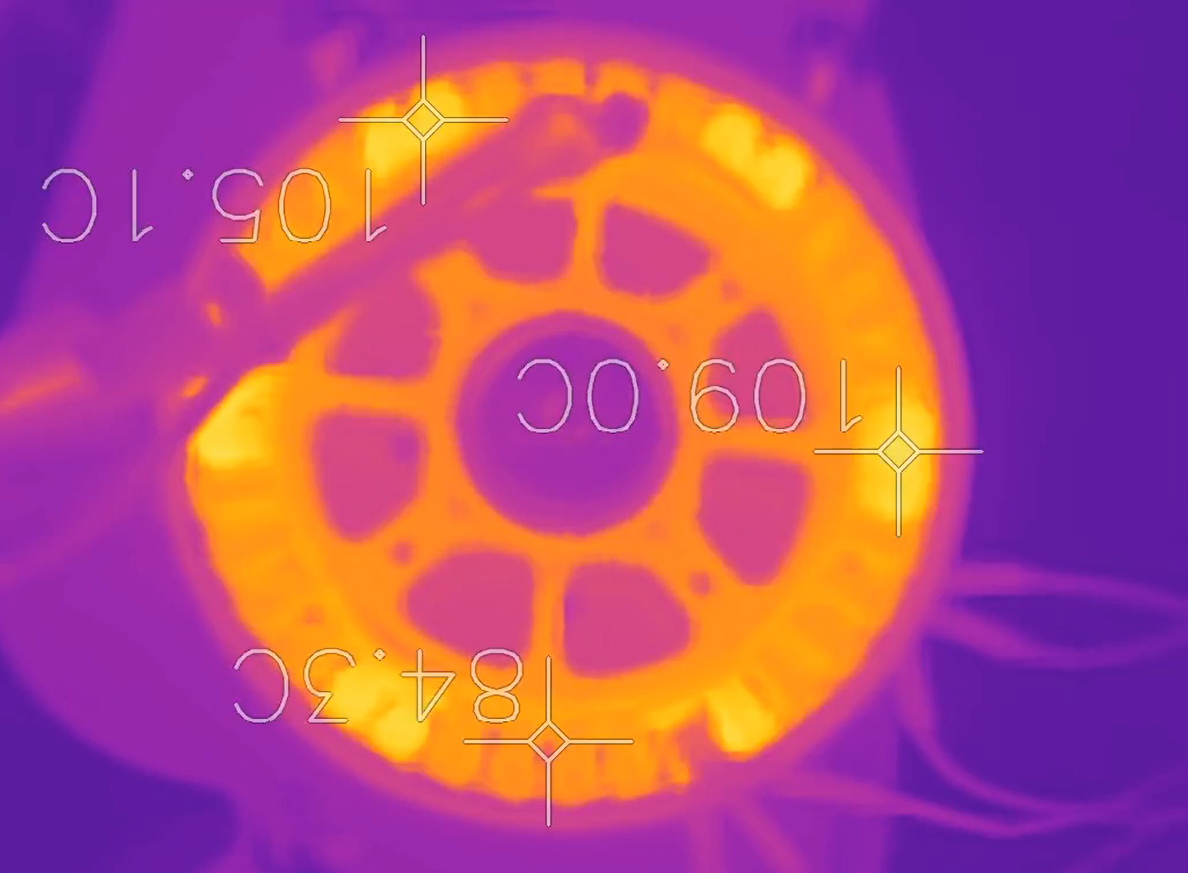}
\caption{A delta wound motor with current flowing through two leads with the majority of the heat produced through one phase. This method can be used to identify winding type by visual inspection.}
\label{figure:thermal_camera}
\end{figure}

Thermal imaging techniques provide a reasonably convenient and economical approach for measuring the surface temperature of objects. The following describes the procedure for identifying the winding configuration of a BLDC motor: 
\begin{enumerate}
    \item Connect two of the leads of the BLDC motor to a DC power supply. We commonly use power supply currents from 4A to 8A.
    \item Direct a thermal imaging camera to the motor.
    \item Allow current to flow through the leads until some windings become visible in the thermal camera, while not overheating the motor excessively.
\end{enumerate}
In the example image (Fig. \ref{figure:thermal_camera}), only one set of windings is producing the majority of the heat while two sets are producing less (4x less resistive loss), indicating that the motor is delta wound.
Wye wound motors would have two sets of windings producing more heat with one set of windings unpowered. 

\section{Motor Physics} \label{sec:appendix_motorphysics}
Here, we briefly review the fundamentals of brushed DC motor operation. Permanent magnet electric motors create force via  interaction between the magnetic field produced from permanent magnets and an electromagnetic field produced by the windings. This force is governed by the Lorentz Force Law. If we consider a one squared-loop of wire in two magnets, the Lorentz Law can be expressed as: 

\begin{equation}
  \bm F = I(\bm \ell  \times \bm B) = I \norm{\bm B}\norm{\bm \ell} \bm n \;\; (\bm \ell \perp \bm B)
\end{equation}

where $\bm F$ is the force created on one side of the wire, $\bm B$ is the magnetic field created by the magnet, $ I$ is current flowing through the wire, and $\bm \ell$ is the length vector of the wire parallel to the rotational axis, and $\bm n$ is a unit vector perpendicular to both wire and the magnetic flux. Then, the torque acting on a whole loop becomes:

\begin{align}
\bm \tau =  2 \bm F \times \bm D =  & 2 \norm{\bm D} \norm{\bm F}\sin({\theta_r}) \, \bm j \\
= & 2 \ \norm{\bm D} \norm{\bm \ell} \norm{ \bm B} I\sin({\theta_r}) \, \bm  j
\end{align}
where $D$ is a half of the width of the loop (Fig. 28-18 of \cite{halliday2013fundamentals}), and $\theta$ is the angle of the loop, and $\bm j$ is a unit vector perpendicular to $\bm F$ and $\bm D$. Please refer to Figure 28-18 of \cite{halliday2013fundamentals} for depiction of the motor.\par
The relative movement of the coil caused by the Lorentz law in the magnetic field induces a voltage (\textit{i.e.} the back-EMF). The back-EMF is generated by the Faraday's Law: 
\begin{align*}
    V_e &= \oint(\bm v \times \bm B) \cdot \bm d\ell = 2\norm{\bm v} \norm{\bm B} \norm{\bm \ell} \\ &= 2\norm{r}\frac{d\theta_r}{dt}\norm{\bm B} \norm{\bm \ell} \;\;\;\; ( \bm v \cross \bm B \parallel \bm \ell)
\end{align*}
where $V_e$, is the induced electromotive force voltage, and $v$ is the tangential velocity of the wire.

\subsubsection{Torque \& Back-EMF production}

The torque acting on the rotor composed of $N$ number of coils is calculated by the Lorentz Law: 
\begin{equation}
    \bm \tau = 2( I BN \ell) D\sin({\theta_r}) \, \bm j 
\end{equation}

For the rotor rotate in a single direction ($\theta>0$), the current must reverse at  $\pm \, 90 \times n$ $(n=1,2,3 ...)$. In a brushed motor, this is enabled by the brushes and physical commutator. Furthermore, by adding the loops of coils and offsetting from each other, a motor can be developed which creates a nearly constant torque (\textit{i.e.} no torque ripple) \cite{franklin2002feedback}. By reflecting these changes, the torque exerted on the rotor can be expressed as:

\begin{equation} \label{eq:torque_brushed}
  \bm \tau = 2(I B N \ell)D \bm j = K_t  I\bm j
\end{equation}
where the torque constant is defined as:
\begin{equation}
K_t = 2B N\ell D
\end{equation}
 
Similarly, the back-EMF of a brushed motor can be calculated:

\begin{equation} \label{eq:emf_brushed}
    V_e = 2(D\frac{d\theta_r}{dt})BN\ell = K_b \frac{d\theta_r}{dt}
\end{equation}
where $K_b$ is the back-EMF constant defined as:
\begin{equation}
K_b  = 2B N\ell D
\end{equation}
These two equations \eqref{eq:torque_brushed}, \eqref{eq:emf_brushed} demonstrate that the torque constant and back-EMF constants are identical:
\begin{equation}
  K_b = K_t  
\end{equation}
Note that this relationship is only valid when using SI units. For different units, adequate conversion should be included in the modeling analysis. The above property stems from the power conversion:
\begin{align}
\tau \frac{d\theta_r}{dt} = I V_e \\
(K_t I) \frac{d\theta_r}{dt} = I (K_b \frac{d\theta_r}{dt}) \\  
K_t \cancel {I \frac{d\theta_r}{dt}} =  K_b \cancel{I \frac{d\theta_r}{dt}}
\end{align}
where it represents the conversion from mechanical to electrical power in an ideal motor. 

\section{Conversion from brushless to single phase ``brushed'' motor quantities (\textit{i.e.} q-axis)} \label{section:appedix_a}

The following section describes a full derivation of the voltage equation that governs BLDC motors. We start from voltage equations of each phase and convert those to the d-q axes. We first introduce the concept of total flux-linkage, which is the total magnetic flux through the windings. The name `flux-linkage' stems from the fact that the windings are linked by the shared magnetic flux \cite{halliday2013fundamentals}. The total flux-linkage is  composed of self and mutual flux-linkages between the stator windings, and the flux-linkages between the permanent magnet and windings (\textit{i.e.} rotor-stator flux-linkage):
\begin{equation}
    \bm{\Psi^\phi} = \bm{L^\phi I^\phi + \Psi^R}
\end{equation}

where, $\bm L^\phi$ and  $\bm I^\phi$ denotes inductances and currents of all phases, and flux-linkages of windings are denoted $\bm{L^\phi I^\phi }$ and flux-linkages between the magnet and windings are denoted $\bm{\Psi^R}$. Each element represents phase quantities of the flux-linkages:
\begin{equation}
\bm \Psi^\phi= 
\begin{bmatrix}
\Psi^\phi_A\\
\Psi^\phi_B\\
\Psi^\phi_C
\end{bmatrix}, 
\bm \Psi^R= 
\begin{bmatrix}
\Psi^R_A\\
\Psi^R_B\\
\Psi^R_C
\end{bmatrix}
\end{equation}

\begin{equation}
\bm I^\phi= 
\begin{bmatrix}
I^\phi_A\\
I^\phi_B\\
I^\phi_C
\end{bmatrix}
\end{equation}

\begin{equation}
\bm L^\phi= 
\begin{bmatrix}
L_s & L_{m} & L_{m} \\
L_{m} & L_s& L_{m} \\
L_{m} & L_{m} & L_s
\end{bmatrix}
\end{equation}
In balanced BLDC motors, the self inductances of each phase and mutual inductances of each pair of phases are identical \cite{niapour2014review}.
To derive the voltage equation of the motor, Kirchoff's Voltage Law can be expressed in matrix form as:
\begin{equation}  \label{eq:voltage_eq_matrix}
    \bm{V^\phi} = \bm{R^\phi I^\phi} + \frac{d \bm \Psi^\phi }{dt}
\end{equation}
where the equation of one representative phase is:
\begin{equation} \label{eq:volt_onephase}
    V^\phi_A = R^\phi I^\phi_A + \frac{d}{dt}(L_s I^\phi_A + L_m I^\phi_B + L_m I^\phi_C) + \frac{d \Psi^R_A}{dt}
\end{equation}
Since we assumed phase currents are sinusoidal and 120$^{\circ}$ out of phase, the phase currents have following property:
\begin{equation}
I^\phi_A + I^\phi_B +I^\phi_C =0
\end{equation}
The above property simplifies the voltage equation (\ref{eq:volt_onephase}) of a single phase to: 
\begin{align}
V^\phi_A &= R^\phi I^\phi_A + (L_s - L_m) \frac{d I^\phi_A}{dt} +  \frac{d \Psi^R_A}{dt} \\
    & = R^\phi I^\phi_A + L^e \frac{d I^\phi_A}{dt} +  \frac{d \Psi^R_A}{dt}
\end{align}
where, we define an effective inductance as:
\begin{equation} \label{eq:effective_induct}
L^e = L_s - L_m
\end{equation}
and back-EMF of a phase as:
\begin{equation}
V_{e,A} = \frac{d \Psi^R_A}{dt} =  K_{b, A}^\phi \frac{d\theta_r}{dt}
\end{equation}
Therefore, applying the same logic on all phases results in following expression:
\begin{equation}
\begin{split}
\begin{bmatrix}
V_A\\
V_B\\
V_C
\end{bmatrix}
= 
\begin{bmatrix}
R^\phi & 0& 0 \\
0 & R^\phi& 0 \\
0 & 0& R^\phi
\end{bmatrix}
\cdot
\begin{bmatrix}
I_A\\
I_B\\
I_C
\end{bmatrix}
+ \\
\begin{bmatrix}
L^e & 0& 0 \\
0 & L^e & 0 \\
0 & 0& L^e
\end{bmatrix}
\cdot
\frac{d}{dt}
\begin{bmatrix}
I_A\\
I_B\\
I_C
\end{bmatrix}
+ 
\begin{bmatrix}
K_{b, A}^\phi\\
K_{b, B}^\phi\\
K_{b, C}^\phi
\end{bmatrix}
\frac{d\theta_r}{dt}
\end{split}
\end{equation}

where the winding voltage and back-EMF constant of each phase is:
\begin{align}
V_A &= -\bar{V^\phi} \sin(\theta)\\
V_B &=- \bar{V^\phi} \sin(\theta-\frac{2}{3}\pi)\\
V_C &=- \bar{V^\phi} \sin(\theta+\frac{2}{3}\pi) 
\end{align}
\begin{align}
K_{b, A}^\phi &= -\bar K_b^\phi \sin(\theta)\\
K_{b, B}^\phi &= -\bar K_b^\phi \sin(\theta-\frac{2}{3}\pi)\\
K_{b, C}^\phi &= -\bar K_b^\phi \sin(\theta+\frac{2}{3}\pi) 
\end{align}
where $V_x, K_{b, x}^\phi$ are phase voltages and phase back-EMF constants, respectively and $\bar{V^\phi}$, $\bar K_b^\phi$ are amplitudes of each phase, which amplitudes are all identical. 

By applying power invariant transformation (\ref{eq:dq_trans}) to the time derivative of the total flux linkage \cite{kirtley2020electric}:

\begin{align}
    \bm {Q} \frac{d}{dt}\bm{\Psi^\phi} & = \bm{Q} \frac{d}{dt} (\bm {Q^{\dagger} \Psi^{dq}}) \\
    & =\bm Q \frac{d}{dt} \bm{Q^{\dagger}}  \bm{\Psi^{dq}} + \frac{d \bm {\Psi^\phi} }{dt}
\end{align}
In here, we denote $\dagger$ as the right inverse ($\bm Q \cdot \bm{ Q^\dagger} = \bm I$). and $\bm Q = \bm P \bm C$ is the d-q transformation \eqref{eq:clarke}, \eqref{eq:park}. By converting all elements in the voltage equation to the d-q axis (\ref{eq:voltage_eq_matrix}): 
\begin{equation}
    \bm{V^{dq}} = \begin{bmatrix}
    V_d \\
    V_q 
    \end{bmatrix}, 
    \bm{\Psi^{dq}} = \begin{bmatrix}
    \Psi_d \\
    \Psi_q 
    \end{bmatrix} 
\end{equation} 
\begin{equation}
    \bm Q \frac{d}{dt} \bm{Q^{\dagger}} = \begin{bmatrix}
    0 & -\frac{d\theta}{dt} \\
    \frac{d\theta}{dt} & 0 
    \end{bmatrix}
\end{equation}
Therefore by substituting above d-q quantities in the BLDC electrical equation, the d-q representation of the voltage equation becomes: 
\begin{align}
    V^d  & = R^\phi I^d + \frac{d\Psi^d}{dt} - \frac{d\theta}{dt} \Psi^q \\
    V^q  & = R^\phi I^q + \frac{d\Psi^q}{dt} + \frac{d\theta}{dt} \Psi^d 
\end{align}
By further expanding the above equation using the flux-linkages:
\begin{equation}
        V^d   =R^\phi I^d + L_d\frac{dI^d}{dt} +\frac{d\bar{\Psi^R}}{dt}- \frac{d\theta}{dt}  L^qI^q
\end{equation}
where, 
\begin{equation}
\Psi_q = L^q I^q
\end{equation}
\begin{equation}
    V^q   = R^\phi I^q + L_q\frac{dI^q}{dt} + \frac{d\theta}{dt} (L^d I^d  + \bar{\Psi^R})
\end{equation}
where,
\begin{equation}
\Psi_d = L^dI^d + \bar{\Psi^R} 
\end{equation}
which demonstrates the magnetic flux-linkage is only in the d-axis. Since the amplitude of the magnetic flux-linkage ($\bar{\Psi^R}$) is constant and the d-q transformation is in phase with the phase currents ($I^d = 0$) (refer to the full list of assumptions in \ref{sec:bldc_modeling}) the d-q axis voltage equation reduces to:
\begin{align}
    V^d  & = R^\phi I^d  - \frac{d\theta}{dt}  L^e I^q \\
    V^q  & = R^\phi I^q + L^e \frac{dI_q}{dt} + \frac{d\theta_r}{dt} K_b^q \tag{\ref{eq:quad_electric}}
\end{align}
where $\bar{\Psi^R} = pK_b^q$, and $L^d=L^q=L^e$ from the d-q transformation as follows \cite{kirtley2020electric}:
\begin{equation}
\bm {L^{dq}} = \bm{QL^\phi Q^{\dagger}} = \begin{bmatrix}
L^d & 0 \\
0 & L^q 
\end{bmatrix}=
\begin{bmatrix}
L^e & 0 \\
0 & L^e
\end{bmatrix}
\end{equation}
The last equality is due to the non-saliency of the rotor. This demonstrates the q-axis voltage equation reduces to a DC representation, and agrees with the characteristics of the surface permanent synchronous motor where reluctance is almost equal in d and q-axis direction (\textit{i.e.} saliency ratio of 1) \cite{meier2002theoretical,bobek2013pmsm}.

 \section{Accurate Power Loss of BLDC Motors} \label{sec:appendix_power_inaccurate}

In the paper, we provided accurate calculation of Joule heating in phase and q-axis quantities, which are agnostic to the winding configuration (Section \ref{sec:power_loss}). However, typically electrical resistance provided in manufacturer datasheets is the terminal resistance, where conversion from terminal to phase resistance requires a knowledge of the motor's winding configuration. In this appendix, we demonstrate how using terminal resistance inappropriately without the adequate use of conversion factors can lead to inaccurate estimation of power loss.

 The following describe the power loss with respect to line-to-line (terminal) quantities on each configuration of the motors. Using (\ref{eq:power_quad}), for wye-wound motors:
\begin{align}
R^{\phi} &=\frac{1}{2} R^{ll}\tag{\ref{eq:resistance_wye}} \\
P &= {I^{q}}^2 R^\phi = \frac{1}{2} {I^{q}}^2 R^{ll} \label{eq:wye_power}
\end{align}
For delta-wound motors: 
\begin{align}
R^{\phi}&=\frac{3}{2} R^{ll}\tag{\ref{eq:resistance_delta}} \\
P &= {I^{q}}^2 R^\phi = \frac{3}{2} {I^{q}}^2 R^{ll}   \label{eq:delta_power}
\end{align}
where $ll$ denotes line-to-line quantities. The power loss (eq. \ref{eq:wye_power}, \ref{eq:delta_power}) shows that if one considers the line-to-line quantities as brushed equivalent quantities (\textit{i.e.} $P = {I^q}^2R^{ll} $), the power loss would be two times higher for wye-wound and 2/3 times lower for delta-wound than the actual power consumption. Thus, when interpreting motor datasheets, the reference frame used for the electrical resistance is important to prevent over or underestimating Joule heating power loss.

\section{Accurate Torque Production of BLDC Motors}
\label{section:appendix_torque_inaccurate}
In this section, we provide more detail on the impact of using mismatched or inconsistent reference frames during analysis. Specifically, in the context of estimating torque production, inaccuracies stem from pairing a torque constant $K_t$ with a current value obtained in different reference frames. As described in Sec. \ref{sec:torque}, a potential error is to consider the back-EMF constant $K_b$ as torque constant $K_t$. Specifically, manufacturers of exterior-rotor-type BLDC motors, also known as ``drone motors,'' typically report only the velocity constant $K_v$ which is the reciprocal of the back-EMF constant $K_b$, where the velocity constant is represented in line-to-line frame (\ref{eq:delta_voltage1}):
\begin{equation}
    \bar{K_v^{ll}} = 1/\bar{K_b^{ll}}
\end{equation}
We assume a delta configuration in this example, where the line-to-line voltage is identical to phase voltage:
\begin{equation}
    \bar{K_b^{ll}} = \bar{K_b^{\phi}} 
\end{equation}
Since phase $K_b$ is equal to phase $K_t$, by using the conversion of $K_t$ described in the main section (\ref{eq:conversion_kt_delta}), the q-axis torque constant becomes:
\begin{equation}
    \bar{K_b^{\phi}} =  \bar{K_t^{\phi}} = \sqrt{\frac{2}{3}}K_t^q
\end{equation}
 Summing up the equations described above, the accurate form of torque estimation using q-axis current becomes:
\begin{equation}
\tau = K_t^q I^q = \sqrt{\frac{3}{2}}\bar{K_b^{\phi}}  I^q  = \sqrt{\frac{3}{2}}\bar{K_b^{ll}}  I^q = \sqrt{\frac{3}{2}} \frac{1}{\bar{K_v^{ll}}}  I^q
\end{equation}

Therefore, incorrectly using the $K_v$ with the q-axis reference frame will lead to $\sqrt{\frac{3}{2}}$ smaller torque estimation than actual. Note that there can be variants of inaccurate estimations depending on which pair of torque and current reference frames are selected, and the demonstrated example is merely one of these potential sources of errors.


\section*{Acknowledgment}

EJR and LMM would like to thank Mr. Rick Casler for inspiring parts of this tutorial.  EJR would like to thank Mr. Nathan Kau for his help reviewing the technical content, and Ms. Hannah Frame for her assistance in creating the figures.

\ifCLASSOPTIONcaptionsoff
  \newpage
\fi


{\small
\bibliographystyle{IEEEtran}
\bibliography{IEEEabrv,main}
}

\end{document}